\documentclass[10pt,twocolumn,letterpaper]{article}

\usepackage{latex8}
\usepackage{times}
\usepackage{graphicx}
\usepackage{amssymb}

\usepackage{amsfonts}
\usepackage{amsmath}
\usepackage{algorithm}
\usepackage{algorithmic}
\usepackage{caption} 
\usepackage{color}
\usepackage{epsfig}
\usepackage{footnote}
\usepackage{multirow}
\usepackage{subcaption}
\usepackage{tablefootnote}

\begin{document}

\title{Probabilistic Attribute Tree in Convolutional Neural Networks for Facial Expression Recognition}

\author{Jie Cai$^1$, Zibo Meng$^2$, Ahmed Shehab Khan$^1$, Zhiyuan Li$^1$, James O'Reilly$^1$, and Yan Tong$^1$\\
\small{$^1$Department of Computer Science \& Engineering, University of South Carolina, Columbia, SC}\\
\small{$^2$Innopeak Technology Inc., Palo Alto, CA}\\
\footnotesize{\{jcai,akhan,zhiyuanl,oreillyj\}@email.sc.edu}, \footnotesize{mzbo1986@gmail.com}, \footnotesize{tongy@cec.sc.edu}
}

\maketitle
\thispagestyle{empty}
\small
\begin{abstract}
In this paper, we proposed a novel Probabilistic Attribute Tree-CNN (PAT-CNN) to explicitly deal with the large intra-class variations caused by identity-related attributes, e.g., age, race, and gender. Specifically, a novel PAT module with an associated PAT loss was proposed to learn features in a hierarchical tree structure organized according to attributes, where the final features are less affected by the attributes. Then, expression-related features are extracted from leaf nodes. Samples are probabilistically assigned to tree nodes at different levels such that expression-related features can be learned from all samples weighted by probabilities. We further proposed a semi-supervised strategy to learn the PAT-CNN from limited attribute-annotated samples to make the best use of available data. Experimental results on five facial expression datasets have demonstrated that the proposed PAT-CNN outperforms the baseline models by explicitly modeling attributes. More impressively, the PAT-CNN using a single model achieves the best performance for faces in the wild on the SFEW dataset, compared with the state-of-the-art methods using an ensemble of hundreds of CNNs.
\end{abstract}

\section{INTRODUCTION}

\begin{figure*}[th]
   \centering
   \includegraphics[width=0.88\textwidth]{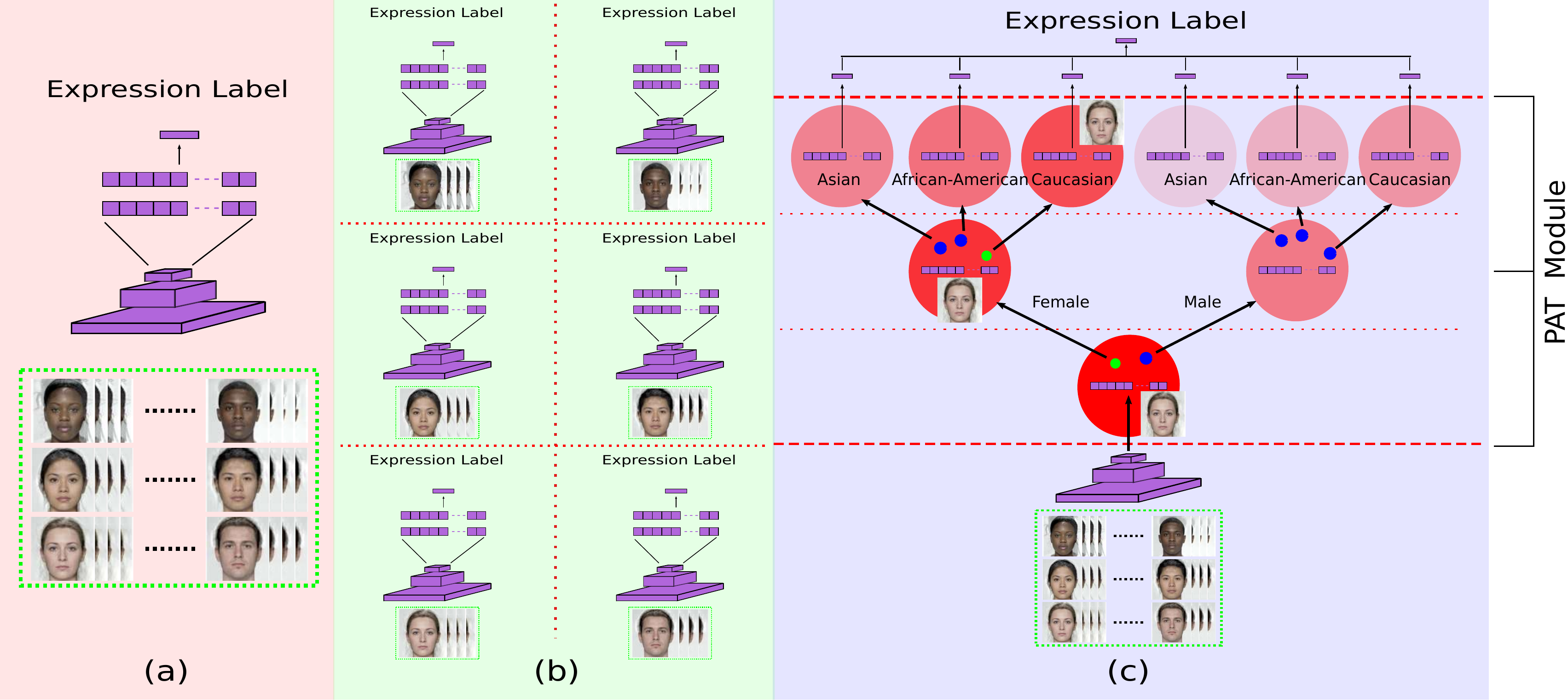}
   \caption{\small{ Deep CNN structures for facial expression recognition: (a) a traditional CNN trained from all data, (b) attribute-specific CNNs trained from subsets of the dataset, and (c) the proposed PAT-CNN, where each node contains an FC layer connected to FC layer in its parent node, if any. Dots represent cluster centers, each of which corresponds to a state of the specific attribute. The green dots denote the cluster centers corresponding to the ground truth attribute states of the current sample, e.g., ``Female'' for gender in the root node. Best viewed in color.}}
   \label{fig:introductionStructure}
\end{figure*}

As one of the most natural, powerful, and universal means of human communication, facial expression has been studied intensively in various active research fields. 
Most recently, deep CNNs have attracted increased attention for facial expression recognition. Despite the CNN structures employed,
most of these approaches were trained to learn expression-related facial features from all subjects as illustrated in Fig.~\ref{fig:introductionStructure} (a), while identity-related attributes, e.g., age, race, and gender, are not explicitly considered.

However, it is widely believed that facial appearance and 3D geometry are determined by person-specific attributes and affected by facial expressions temporarily. For instance, it is hard to differentiate the transient wrinkles caused by facial expressions from the permanent ones of elder adults. In addition, the presence of facial hair such as beards may introduce various occlusions for male versus female subjects.

Furthermore, studies in psychology have shown that emotional expressions demonstrate considerable differences across age~\cite{gross1997emotion}, race~\cite{vrana2002role}, and gender~\cite{vrana2002role,hess2004facial,simon2004gender} in terms of expression intensity. For examples, Asian people show consistently lower intensity-expressions than the other ethnic groups~\cite{matsumoto1993ethnic}; and women were found to express more anger~\cite{hess2004facial} and sadness~\cite{simon2004gender} than men do.

Due to high inter-subject variations caused by attributes, it remains challenging to learn expression-related features with CNNs, especially from static images. This motivates us to alleviate identity-related variations by explicitly modeling person-specific attributes in CNNs.

An intuitive solution is to train multiple attribute-specific CNNs from subsets of the dataset. As shown in Fig.~\ref{fig:introductionStructure} (b), a set of CNNs can be trained from different combinations of attributes, respectively. However, unlike large-scale datasets for object detection or categorization, expression-labeled datasets are much smaller and most of them lack attribute annotations. Moreover, classifying attributes in the real world is still a challenging problem. Therefore, recognition performance of attribute-specific CNNs is likely to degrade due to misclassified attributes and insufficient training data in the subsets as demonstrated in our experiments.

In this work, we proposed a novel PAT-CNN, where features are learned through a hierarchical tree structure organized according to attributes. As shown in Fig.~\ref{fig:introductionStructure} (c), a novel PAT module is embedded right after the last pooling layer. The PAT module has a hierarchical tree structure, where each node contains a fully-connected (FC) layer connected to the FC layer in its parent node, if any.

Given a set of training samples, clustering is conducted at each node except the leaf nodes according to a type of attributes by using features from the current FC layer in that node. Hence, the number of child nodes is determined by the number of clusters, i.e., the number of states of the attribute. For example, as shown in Fig.~\ref{fig:introductionStructure} (c), clustering is performed in terms of ``Gender'' at the root node and results in two clusters corresponding to two child nodes for male and female, respectively. Since the goal is NOT attribute recognition, each data sample is probabilistically assigned to all nodes. As depicted in Fig.~\ref{fig:introductionStructure} (c), a female Caucasian has a high probability to be assigned to ``Female'' node at the second level and ``Female Caucasian'' node at the third level. Expression-related features, i.e., the output of the PAT module, are extracted from the FC layers at the leaf nodes, from which a set of expression classifiers are trained. The final decision of expression classification is achieved by a weighted sum of all expression classifiers.


Furthermore, a semi-supervised learning strategy is developed to learn the PAT-CNN from limited attribute-annotated data. In addition to the loss for expression classification, a novel PAT loss function is developed to iteratively update cluster centers, shown as blue and green dots in Fig.~\ref{fig:introductionStructure} (c), during training, which ensures clustering results are semantically meaningful. Specifically, the data samples with attribute labels are used to minimize the PAT loss; whereas all samples with expression labels are employed to minimize the expression loss. \emph{Note that the attribute labels are only used in training, but not in testing.}


In summary, our major contributions are:
\vspace{-0.06in}
\begin{enumerate}
\item[-] Developing a PAT-CNN to alleviate variations introduced by person-specific attributes for facial expression recognition.
\vspace{-0.08in}
\item[-] Developing a novel PAT module with an associated PAT loss to learn expression-related features in a hierarchical manner, where the output features of the PAT module are less affected by attributes.
\vspace{-0.08in}
\item[-] Developing a semi-supervised learning strategy to train the PAT-CNN from limited attribute-annotated data, making the best use of available facial expression datasets.
\end{enumerate}
\vspace{-0.08in}

Extensive experiments on five expression datasets show that the proposed PAT-CNN yields considerable improvement over the baseline CNNs learned from all training data (Fig.~\ref{fig:introductionStructure} a) as well as the attribute-specific CNNs learned from subsets of the dataset (Fig.~\ref{fig:introductionStructure} b). We also showed that the proposed soft-clustering with probability outperforms the one based on hard-clustering using the same network structure. More impressively, the PAT-CNN using a single model achieves the best performance for faces in the wild on the SFEW dataset, compared with state-of-the-art methods using an ensemble of hundreds of CNNs.

\section{RELATED WORK}
Facial expression recognition has been extensively studied as elaborated in the recent surveys~\cite{Sariyanidi2015,Martinez2017Automatic}. One of the major steps in facial expression recognition is to extract features that capture the appearance and geometry changes caused by facial behavior, from either static images or dynamic sequences. These features can be roughly divided into two main categories: hand-crafted and learned features. Recently, features learned by deep CNNs have achieved promising results, especially in more challenging settings. Most of these approaches were trained from all training data, whereas attribute-related and expression-related facial appearances are intertwined in the learned features. While progress has been achieved in choices of features and classifiers, the challenge posed by subject variations remains for person-independent recognition.

More recently, identity information is explicitly taken into consideration when learning the deep models. An identity-aware CNN~\cite{meng2017identity} developed an identity-sensitive contrastive loss to learn identity-related features. An Identity-Adaptive Generation (IA-gen) method~\cite{yang2018identity} was proposed to synthesize person-dependent facial expressions from any input facial images using six conditional Generative Adversarial Networks (cGANs); and then recognition is performed by comparing the query images and the six generated expression images, which share the same identity information. The cGAN was also used in De-expression Residue Learning (DeRL) ~\cite{yang2018facial} to generate a neutral face image from any input image of the same identity, while the residue of the generative model contains person-independent expression information.

Apart from learning identity-free expression-related features, Multi-task Learning (MTL) has been employed~\cite{ranjan2017all} to simultaneously perform various face-related tasks including detection, alignment, pose estimation, gender recognition, age estimation, smile detection, and face recognition using a single deep CNN. To deal with the incomplete annotations and thus, insufficient and unbalanced training data for various tasks, the all-in-one framework was split into subnetworks, which were trained individually. Our approach differs significantly from the MTL~\cite{ranjan2017all} that we jointly minimize the loss of the major task, i.e., expression recognition errors, and those of the auxiliary tasks, i.e., the PAT loss, calculated in a hierarchical tree structure. In addition, semi-supervised learning is employed in our approach to make the best use of all available data.

Recently, clustering has been utilized to group deep features. A recurrent framework~\cite{yang2016joint} updates deep features and image clusters alternatively until the number of clusters reaches the predefined value. DeepCluster~\cite{caron2018deep} alternatively groups the features by k-means and uses the subsequent assignments as supervision to learn the network. Deep Density Clustering (DDC)~\cite{lin2018deep} groups unconstrained face images based on local compact representations and a density-based similarity measure. In contrast to these unsupervised clustering methods, the proposed PAT-CNN takes advantage of available attribute annotations and thus, is capable of learning semantically-meaningful clusters that are related to facial expression recognition. Moreover, data samples are probabilistically assigned to clusters at different levels of the hierarchy to alleviate the misclassifications due to clustering errors.

\section{METHODOLOGY}

\vspace{-0.13in}
In this section, we will first introduce the proposed PAT module, then present the PAT loss and the corresponding forward and backward propagation processes. Finally, we will show the overall loss function of the PAT-CNN.

\subsection{The Overview of the PAT Module}

\begin{figure}[th]
   \centering
   \includegraphics[width=0.48\textwidth]{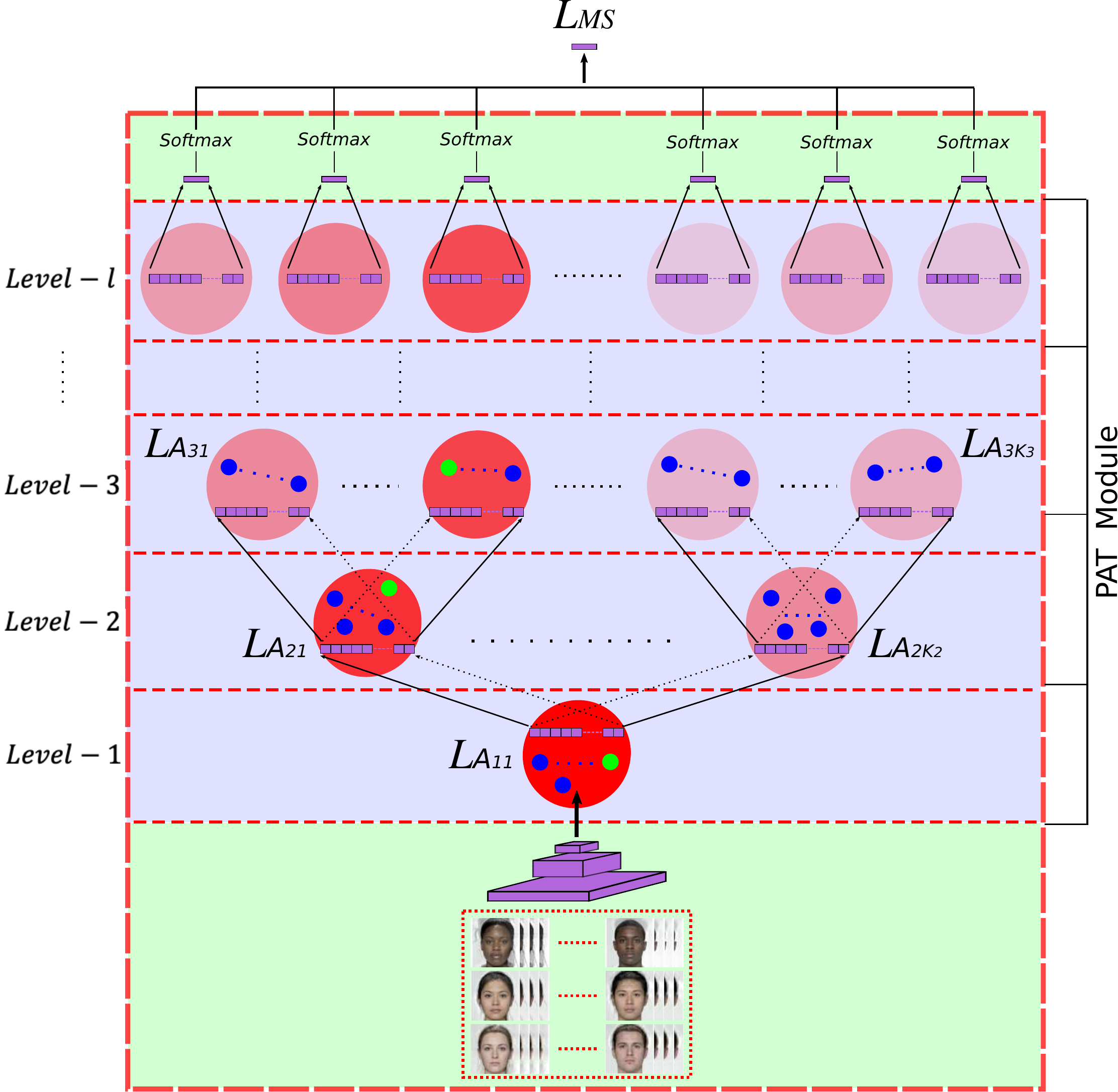}
   \caption{\small{An illustration of the PAT-CNN. Each node contains an FC layer, which connects to another FC layer located at its parent node, if any. At each node except the leaf nodes, features are extracted at the associated FC layer and clustered according to a specified attributes, e.g., age , gender, and race, with centers marked by dots. The green dots denote the cluster centers corresponding to the ground truth attribute states of the current sample. Intensity of red nodes represents the probability of the current sample belonging to the node. Best viewed in color.}}
   \label{fig:clusteringLayer}
\end{figure}

The architecture of the proposed PAT-CNN is illustrated in Fig.~\ref{fig:clusteringLayer}, where a general \emph{l}-Level PAT module is embedded between the last pooling layer and the decision layer of the CNN. Level-1 of the PAT contains the root node, which has one FC layer connected to the last pooling layer. Starting from Level-2, each level consists of a number of nodes, each of which contains an FC layer connected to the FC layer located at its parent node at the previous level.

Excepting the leaf nodes, clustering is performed at each PAT node according to a type of attribute, e.g., age, race, and gender; and nodes in the same level consider the same type of attribute. As shown in Fig.~\ref{fig:clusteringLayer}, features extracted from the associated FC layer are clustered into a number of clusters with the centers denoted as blue or green dots. The number of clusters is determined by the number of states of the specific attribute, e.g., 2 for gender. Furthermore, these clusters also correspond to its child nodes in the next level. As shown in Fig.~\ref{fig:introductionStructure} (c), a ``Female'' cluster in the root node corresponds to one of its child nodes, i.e., the ``Female'' node. 

During training, the samples with the attribute labels will be used to update the cluster centers and learn the parameters of the FC layers of the PAT module by minimizing the proposed PAT loss. As shown in Fig.~\ref{fig:clusteringLayer}, each sample contributes to all nodes differently according to its probabilities of belonging to the nodes illustrated by the color intensities of the nodes. Specifically, a sample contributes more to those nodes containing the green dots, which denote the cluster centers corresponding to its ground truth attribute states, but also to other nodes with lower probabilities. As shown in Fig.~\ref{fig:introductionStructure} (c), a sample of a female Caucasian will contribute more to learn the parameters associated with the ``Female'' node at the second level and ``Female Caucasian'' node at the third level, but less to the other nodes.

\subsection{The PAT Loss: Forward Propagation}

The PAT loss denoted as $\mathcal{L}_{_{P\!A\!T}}$ measures how far away the data samples are from their corresponding cluster centers and is calculated from samples with attribute labels. It is defined as the summation of the attribute losses of all levels except the leaf one:
\begin{footnotesize}
\begin{equation} \label{eq:PAT_loss_forward}
\mathcal{L}_{_{P\!A\!T}}=\sum\limits_{j=1}^{l-1} \mathcal{L}_{A_j}
\end{equation}
\end{footnotesize}
where $l$ is the number of PAT levels. $\mathcal{L}_{A_j}$ is the attribute loss of the $j^{th}$ level, which is defined as:\vspace{-0.1in}
\begin{footnotesize}
\begin{equation} \label{eq:PAT_loss_level_forward}
\mathcal{L}_{A_j} =  \sum\limits_{k=1}^{K_j} \mathcal{L}_{A_{jk}}
\end{equation}
\end{footnotesize}
where $K_j$ is the number of tree nodes at the $j^{th}$ level; and $\mathcal{L}_{A_{jk}}$ is the attribute loss of the $k^{th}$ node at the $j^{th}$ level. From now on, the subscript $j$ and $k$ denote the variable at the $j^{th}$ level and the $k^{th}$ node, respectively.

Let $\mathcal{C}_{jk}$ denote a set of cluster centers and $\textbf{x} _{ijk}$ denote the feature vector of the $i^{th}$ sample extracted from the FC layer in the $k^{th}$ node at the $j^{th}$ level. Given $N_{a_j}$ data samples with the attribute labels, $\mathcal{L}_{A_{jk}}$ is calculated as:
\begin{scriptsize}
\begin{equation} \label{eq:PAT_loss_level_node_forward}
\mathcal{L}_{A_{jk}} \!= \!\left \{\!\!\!\!
\begin{array}{c c}

\begin{aligned}
\sum\limits_{i=1}^{N_{a_j}}  & \sum\limits_{ \substack{\textbf{c}_{jkm} \in \mathcal{C}_{jk} \\ ~\textbf{c}_{jkm} \neq \textbf{c}_{y^{a}_{ij}} }} p\left( \textbf{c}_{jkm} |  \textbf{x} _{ijk} \right) \left(1 + D(\textbf{x} _{ijk},\textbf{c}_{jkm})  \right)
\\ &
+\sum\limits_{i=1}^{N_{a_j}} p\left(  \textbf{c} _{y^{a}_{ij}} | \textbf{x} _{ijk} \right) \left( 1 - D(\textbf{x} _{ijk},\textbf{c}_{y^{a}_{ij}})  \right)
,
\end{aligned}

& \!\!\!\!\textbf{c}_{y^{a}_{ij}} \in \mathcal{C}_{jk} \\

\sum\limits_{i=1}^{N_{a_j}} \sum\limits_{ \textbf{c}_{jkm} \in \mathcal{C}_{jk} } p\left( \textbf{c}_{jkm} |  \textbf{x} _{ijk} \right) \left(1 + D(\textbf{x} _{ijk},\textbf{c}_{jkm})  \right)    , & \!\!\!\!\text{otherwise}

\end{array}
\right.
\end{equation}
\end{scriptsize}
where $y^{a}_{ij}$ denotes the ground truth attribute label of the $i^{th}$ sample at the $j^{th}$ level, e.g., gender =``Female'' in Fig.~\ref{fig:introductionStructure} (c). $\textbf{c}_{jkm} \in \mathcal{C}_{jk}$ denotes the $m^{th}$ cluster center in the $k^{th}$ node, i.e., a dot in the $k^{th}$ node; and $\textbf{c} _{y^{a}_{ij}} \in \mathcal{R}^{d} $ denotes the cluster center corresponding to the ground truth attribute label, i.e., the only green dot at the $j^{th}$ level, in Fig.~\ref{fig:introductionStructure} (c) and Fig.~\ref{fig:clusteringLayer}. $D(\textbf{a},\textbf{b})$ is a distance function and defined as a cosine distance between two vectors in this work.


$\textbf{c}_{jkm}$ has three conditions: (1) $\textbf{c}_{jkm} =\textbf{c} _{y^{a}_{ij}}$ (the $m^{th}$ cluster is denoted by a green dot), (2) $\textbf{c}_{y^{a}_{ij}} \in \mathcal{C}_{jk}$ and $\textbf{c}_{jkm} \neq\textbf{c} _{y^{a}_{ij}}$ (the $k^{th}$ node contains the green dot, but the $m^{th}$ cluster is denoted by a blue dot); and (3) $\textbf{c}_{y^{a}_{ij}} \notin \mathcal{C}_{jk}$ (the $k^{th}$ node contains all blue dots).


Thus, each attribute loss $\mathcal{L}_{A_{jk}}$ is calculated by Eq.~\ref{eq:PAT_loss_level_node_forward} in two cases. Using Fig.~\ref{fig:introductionStructure} (c) as an example, both ``Female'' and ``Male'' nodes contain three clusters according to ``Race''. In the first case, $\mathcal{L}_{A_{jk}}$ of the ``Female'' node is calculated such that the current sample, i.e., a female Caucasian, will be pushed to the center ``Caucasian'' by minimizing the loss term $\left( 1 - D(\textbf{x} _{ijk},\textbf{c}_{y^{a}_{ij}})  \right)$ and pulled away from the other centers by minimizing the loss term $\left(1 + D(\textbf{x} _{ijk},\textbf{c}_{jkm})  \right)$. In the second case, $\mathcal{L}_{A_{jk}}$ of the ``Male'' node is calculated such that the sample will be pulled away from all the centers by minimizing the loss $\left(1 + D(\textbf{x} _{ijk},\textbf{c}_{jkm})  \right)$.

$p\left( \textbf{c}_{jkm} | \textbf{x} _{ijk}\right)$ is the probability of $\textbf{x} _{ijk}$ belonging to the $m^{th}$ cluster and also \emph{the probability of the $i^{th}$ sample being assigned to the $m^{th}$ child node at the $j+1^{th}$ level.} $p\left( \textbf{c}_{jkm} | \textbf{x} _{ijk}\right)$ is defined as:

\begin{scriptsize}
\begin{equation} \label{eq:p(x,c)_function}
p\left( \textbf{c}_{jkm} | \textbf{x} _{ijk} \right)  = \left \{
\begin{array}{c c}
\frac{exp\left( D(\textbf{c}_{jkm},\textbf{x}_{ijk})
\right)}{ \sum\limits_{ \substack{\textbf{c}_{jkn} \in \mathcal{C}_{jk}}}  exp\left( D(\textbf{c}_{jkn},\textbf{x}_{ijk})\right) },

& j=1 \\

q_{ijk}*\frac{exp\left( D(\textbf{c}_{jkm},\textbf{x}_{ijk})\right) }{ \sum\limits_{ \substack{\textbf{c}_{jkn} \in \mathcal{C}_{jk}}}  exp\left(D(\textbf{c}_{jkn},\textbf{x}_{ijk})\right)}  , & \text{otherwise}

\end{array}
\right.
\end{equation}
\end{scriptsize}
where $q_{ijk}$ is the probability of the $i^{th}$ sample belonging to the $k^{th}$ node at the $j^{th}$ level and is calculated at its parent node as described above.

\subsection{The PAT Loss: Backward Propagation}
The partial derivative of the PAT loss $\mathcal{L}_{_{P\!A\!T}}$ with respect to the input sample can be calculated at each node as:

\begin{scriptsize}
\begin{equation} \label{eq:PAT_loss_level_node_backward}
\frac{\partial \mathcal{L}_{A_{jk}}}{\partial \textbf{x} _{ijk} }\!\!\! =\!\!\! \left \{\!\!\!\!
\begin{array}{c c}

\begin{aligned}
\frac{1}{|\mathcal{C}_{jk}|-1} & \sum\limits_{ \substack{\textbf{c}_{jkm} \in \mathcal{C}_{jk} \\ ~\textbf{c}_{jkm} \neq \textbf{c}_{y^{a}_{ij}} }} \!\!p\left( \textbf{c}_{jkm} | \textbf{x} _{ijk} \right) \frac{\partial D(\textbf{x} _{ijk},\textbf{c}_{jkm}) }{\partial \textbf{x} _{ijk} }

 \\ &
 -p\left(  \textbf{c} _{y^{a}_{ij}} | \textbf{x} _{ijk} \right) \frac{\partial D(\textbf{x} _{ijk},\textbf{c}_{y^{a}_{ij}}) }{\partial \textbf{x} _{ijk} },
\end{aligned}
 & \textbf{c}_{y^{a}_{ij}} \in \mathcal{C}_{jk} \\

\frac{1}{|\mathcal{C}_{jk}|}\sum\limits_{ \textbf{c}_{jkm} \in \mathcal{C}_{jk} } p\left( \textbf{c}_{jkm} | \textbf{x} _{ijk} \right) \frac{\partial D(\textbf{x} _{ik},\textbf{c}_{jkm}) }{\partial \textbf{x} _{ijk} }   ,   & \text{otherwise}

\end{array}
\right.
\end{equation}
\end{scriptsize}
where $|\mathcal{C}_{jk}|$ is the number of cluster centers in the $k^{th}$ node and $\frac{\partial D(\textbf{x} _{ijk},\textbf{c}_{jkm}) }{\partial \textbf{x} _{ijk} } $ is defined as:
\begin{scriptsize}
\begin{equation} \label{eq:p(x,c)_function_with_x}
\frac{\partial D(\textbf{x} _{ijk},\textbf{c}_{jkm}) }{\partial \textbf{x} _{ijk} } = \frac{\textbf{c}_{jkm}}{ \|\textbf{c}_{jkm}\|_{_2} \|\textbf{x}_{ijk}\|_{_2} } - \left( \frac{\textbf{c}_{jkm} \cdot \textbf{x}_{ijk}}{ \|\textbf{c}_{jkm}\|_{_2} \|\textbf{x}_{ijk}\|_{_2}^{3} } \right) \textbf{x}_{ijk}
\end{equation}
\end{scriptsize}


Then, the $m^{th}$ cluster center can be updated iteratively in each mini-batch with a learning rate $\alpha$:
\begin{scriptsize}
\begin{equation} \label{eq:update_c}
\textbf{c}_{jkm}^{t+1} = \textbf{c}_{jkm}^{t} - \alpha  \Delta \textbf{c}_{jkm}^{t}
\end{equation}
\end{scriptsize}
where
\begin{scriptsize}
\begin{equation}    \label{eq:PAT_loss_level_node_backward_c}
 \begin{aligned}
\Delta  \textbf{c}_{jkm}=  &  \frac{\sum\limits_{i=1}^{N_{a_j}} -\delta(y^{a}_{ij},m) p\left( \textbf{c}_{jkm} | \textbf{x} _{ijk} \right)
\frac{\partial D(\textbf{x} _{ijk},\textbf{c}_{jkm}) }{\partial \textbf{c}_{jkm} } } {1+\sum\limits_{i=1}^{N_{a_j}}\delta(y^{a}_{ij},m)}
+
\\ &
 \frac{\sum\limits_{i=1}^{N_{a_j}} \sigma (y^{a}_{ij},m) p\left( \textbf{c}_{jkm} |  \textbf{x} _{ijk} \right) \frac{\partial D(\textbf{x} _{ijk},\textbf{c}_{jkm}) }{\partial \textbf{c}_{jkm} } }
{1+\sum\limits_{i=1}^{N_{a_j}}\sigma (y^{a}_{ij},m)}
 \end{aligned}
\end{equation}
\end{scriptsize}
\begin{scriptsize}
\begin{equation} \label{eq:p(x,c)_function_with_c}
\frac{\partial D(\textbf{x} _{ijk},\textbf{c}_{jkm}) }{\partial \textbf{c}_{jkm} } = \frac{\textbf{x}_{ijk}}{ \|\textbf{x}_{ijk}\|_{_2} \|\textbf{c}_{jkm}\|_{_2} } - \left( \frac{\textbf{x}_{ijk} \cdot \textbf{c}_{jkm}}{ \|\textbf{x}_{ijk}\|_{_2} \|\textbf{c}_{jkm}\|_{_2}^{3} } \right) \textbf{c}_{jkm}
\end{equation}
\end{scriptsize}
%
%

\begin{scriptsize}
\begin{equation}
\begin{array}{cc}

\delta(y^{a}_{ij},m) =\left \{
\begin{array}{cc}
1, &  y^{a}_{ij}=m   \\
0, &  \text{otherwise}
\end{array}
\right. &
\sigma(y^{a}_{ij},m) =\left \{
\begin{array}{cc}
0, & y^{a}_{ij}=m           \\
1, & \text{otherwise}
\end{array}
\right.

\end{array}
\end{equation}
\end{scriptsize}

\subsection{A Marginal Softmax Loss}

As shown in Fig.~\ref{fig:clusteringLayer}, the outputs of the PAT module, i.e., features extracted from the FC layers at the leaf nodes in the $l^{th}$ level, are used to train a set of expression classifiers, respectively. The final expression decision is achieved by a weighted sum of all expression classifiers. Given $N_e$ data samples with expression labels, the marginal softmax loss function is defined as follows:
\begin{scriptsize}
\begin{equation} \label{eq:marginal_softmax_loss}
 \begin{aligned}
\mathcal{L}_{MS} =-\sum\limits_{i=1}^{N_{e}} log \sum\limits_{k=1}^{K_{l}} p(y^{e}_{i}|\textbf{x} _{ilk}) q_{ilk}
 \end{aligned}
\end{equation}\end{scriptsize}
where $\textbf{x} _{ilk}$ is the feature vector of the $i^{th}$ sample extracted from the $k^{th}$ node, i.e., the leaf level, of PAT; $K_{l}$ is the number of leaf nodes at the $l^{th}$ level and is also the number of expression classifiers. $p(y^{e}_{i}|\textbf{x} _{ilk})$ is the prediction score of the $i^{th}$ sample displaying the expression of ${y^{e}_{i}}$. $q_{ilk}$ is the probability of the $i^{th}$ sample belonging to the $k^{th}$ node at the $l^{th}$ level and propagated from its parent node, as described previously.


Therefore, given $N_a$ data samples with attribute labels and $N_e$ data samples with expression labels, the overall loss function of the PAT-CNN training is given below:
\begin{footnotesize}
\begin{equation}
\label{eq:joint_hierarchical_loss}
\mathcal{L}=\mathcal{L}_{MS}+\lambda \mathcal{L}_{_{P\!A\!T}}
\end{equation}\end{footnotesize}
where $\lambda$ is a hyperparameter~\footnote{In our experiments, we set $\alpha=1$, $\lambda=0.1$ empirically.} to balance the two losses.

Note that $N_a$ and $N_e$ are not necessarily the same and $\mathcal{L}_{_{P\!A\!T}}$ can be calculated from a small subset of attribute-labeled data. This enables a semi-supervised learning of the PAT-CNN and makes it feasible to improve expression recognition for those existing datasets without attribute labels with the help of additional attribute-labeled data.

The forward and backward training process in the PAT-CNN is summarized in Algorithm~\ref{algo:algo}.

\footnotesize
\begin{algorithm}[th!]
\caption{Forward-backward learning algorithm of PAT-CNN}
\renewcommand{\algorithmicrequire}{\textbf{Input:}}
\renewcommand{\algorithmicensure}{\textbf{Output:}}
\renewcommand{\algorithmiccomment}{\textbf{ //}}
\begin{algorithmic}[1] \label{algo:algo}
    \REQUIRE{Training data $\{\textbf{x}_{i}\}$. }
    \STATE{\textbf{Given:}  number of iterations $T$, learning rates $\mu$ and $\alpha$, and hyperparameter $\lambda$.}
    \STATE{\textbf{Initialize:} $t=1$, network layer parameters $\mathcal{W}$, marginal softmax loss parameters $\theta$, and PAT loss parameters, i.e., the initial cluster centers.}
    \FOR{ $t=1$ to $T$ }
        	\STATE{Calculate the joint loss as in Eq.~\ref{eq:joint_hierarchical_loss}: }
    		\STATE{\quad  \quad  $\mathcal{L}=\mathcal{L}_{MS}+\lambda  \mathcal{L}_{_{P\!A\!T}}$}

    		\STATE{Update the marginal softmax loss parameters: }
    		\STATE{\quad  \quad   $\theta^{t+1}=\theta^{t}-\mu \frac{\partial \mathcal{L}_{MS}^{t}}{\partial \theta ^{t}}$}
    		
    		\STATE{Update the PAT loss parameters (i.e. cluster centers) as in Eq.~\ref{eq:update_c}: }
    		\STATE{\quad   \quad  $\textbf{c}_{k}^{t+1} = \textbf{c}_{k}^{t} - \alpha  \Delta \textbf{c}_{k}^{t} $}

    		\STATE{Update the backpropagation error: }
    		\STATE{\quad  \quad $\frac{\partial \mathcal{L}^{t}}{\partial \textbf{x}_{i}^{t}}=\frac{\partial \mathcal{L}_{MS}^{t}}{\partial \textbf{x}_{i}^{t}}+\lambda  \frac{\partial \mathcal{L}_{_{P\!A\!T}}^{t}}{\partial \textbf{x}_{i}^{t}}$}
    		
    	    \STATE{Update the network layer parameters: }
    	    \STATE{\quad   \quad $ \mathcal{W}^{t+1}=\mathcal{W}^{t} - \mu \frac{ \partial \mathcal{L}^{t} }{ \partial \mathcal{W}^{t} } = \mathcal{W}^{t} - \mu  \frac{\partial \mathcal{L}^{t}}{\partial \textbf{x}_{i}^{t}} \frac{\partial \textbf{x}_{i}^{t}}{\partial \mathcal{W}^{t}}  $}

    \ENDFOR
    \ENSURE{Network layer parameters $\mathcal{W}$, marginal softmax loss parameters $\theta$, and PAT loss parameters, i.e., the cluster centers at the level $1$ to $l-1$.}
 \end{algorithmic}
\end{algorithm}
\normalsize

\section{EXPERIMENTS}
To evaluate the proposed PAT-CNN, experiments have been conducted on five benchmark datasets including three posed facial expression datasets, i.e., the BU-3DFE dataset~\cite{yin20063d}, the CK+ dataset~\cite{Kanade2000,Lucey2010}, and the MMI dataset~\cite{pantic2005web}, and more importantly, two spontaneous ones, i.e., the Static Facial Expression in the Wild (SFEW) dataset~\cite{dhall2015video} and the RAF-DB dataset~\cite{li2017reliable}.

\subsection{Preprocessing}
Face alignment was employed on each image based on centers of two eyes and nose, extracted by Discriminative Response Map Fitting (DRMF)~\cite{asthana2013robust}.
The aligned facial images were then resized to $256\times 256$.
In addition, histogram equalization was utilized to improve the contrast in facial images.
For data augmentation purpose, $224\times 224$ patches were randomly cropped from the $256\times 256$ images, and then rotated by a random degree between -5$^\circ$ and 5$^\circ$. Finally, the rotated images were randomly horizontally flipped as the input of all CNNs in comparison.

\subsection{Experimental Datasets}

\noindent  \emph{\textbf{RAF-DB dataset}} provides attribute labels, e.g., age, race, and gender, for each image and thus, was employed as the major dataset for experimental validation in this work.
Specifically, the single-label subset of the RAF-DB (12,271 training images and 3,068 testing images) was employed, where each image was labeled as one of seven expressions, i.e., neutral and six basic expressions. The subjects were divided into five age groups: (1) 0-3, (2) 4-19, (3) 20-39, (4) 40-69, (5) $70+$ with the gender attribute labeled as one of the three categories, i.e., male, female, and unknown~\footnote{In this work, we discarded 752 training images with gender attribute labeled as unknown.}, and the race attribute labeled as one of the three categories, i.e., African-American, Asian, and Caucasian.

\noindent  \emph{\textbf{SFEW dataset}} is the most widely used benchmark dataset for facial expression recognition in the wild. It contains 1,766 facial images and has been divided into three sets, i.e. Train (958), Val (436), and Test (372). Each image has one of seven expression labels, i.e., neutral and six basic expressions. The expression labels of the ``Test'' set are not publicly available. Thus, the performance on the ``Test'' set was evaluated and provided by the challenge organizer.

\noindent  \emph{\textbf{BU-3DFE dataset}} consists of 2,500 pairs of static 3D face models and 2D texture images from 100 subjects with a variety of ages and races. Each subject displays six basic expressions with four levels of intensity and a neutral expression. Following~\cite{yang2018facial,yang2018identity}, we employed only the 2D texture images of the six basic expressions with high intensity (i.e., the last two levels) in our experiment. Thus, an experimental dataset including 1,200 images was built for the BU-3DFE dataset.


\noindent  \emph{\textbf{CK+ dataset}} consists of 327 videos collected from 118 subjects, each of which was labeled with one of seven expressions, i.e., contempt and six basic expressions. Each video starts with a neutral face, and reaches the peak in the last frame. To collect more data, the last three frames
were collected as peak frames associated with the provided label. Hence, an experimental dataset including 981 images was built for seven expressions and 927 images for six basic expressions.

\noindent  \emph{\textbf{MMI dataset}} contains 236 sequences from 32 subjects, from which 208 sequences of 31 subjects display six basic expressions captured in frontal-view were normally used in experimental validation.
Sequences in MMI start from a neutral expression, through a peak phase near the middle, and back to a neutral face at the end.
Since the actual location of the peak frame was not provided, three frames in the middle of each sequence were collected with the labeled expression
Thus, a total of 624 images were used in our experiments.

\subsection{Training/testing strategy}
Because the three posed facial expression datasets do not provide specified training, validation, and test sets, a person-independent 10-fold cross-validation strategy was employed, where each dataset was further split into 10 subsets, and the subjects in any two subsets are mutually exclusive. For each run, data from 8 sets were used for training, the remaining two subsets were used for validation and testing, respectively. The results were reported as the average of the 10 runs on the testing sets. For the experiments on the SFEW and RAF-DB datasets, we used their training sets for training and their validation and/or testing sets for evaluation, respectively.

\subsection{CNN Implementation Details}
\begin{figure}[th]
   \centering
   \includegraphics[width=0.45\textwidth]{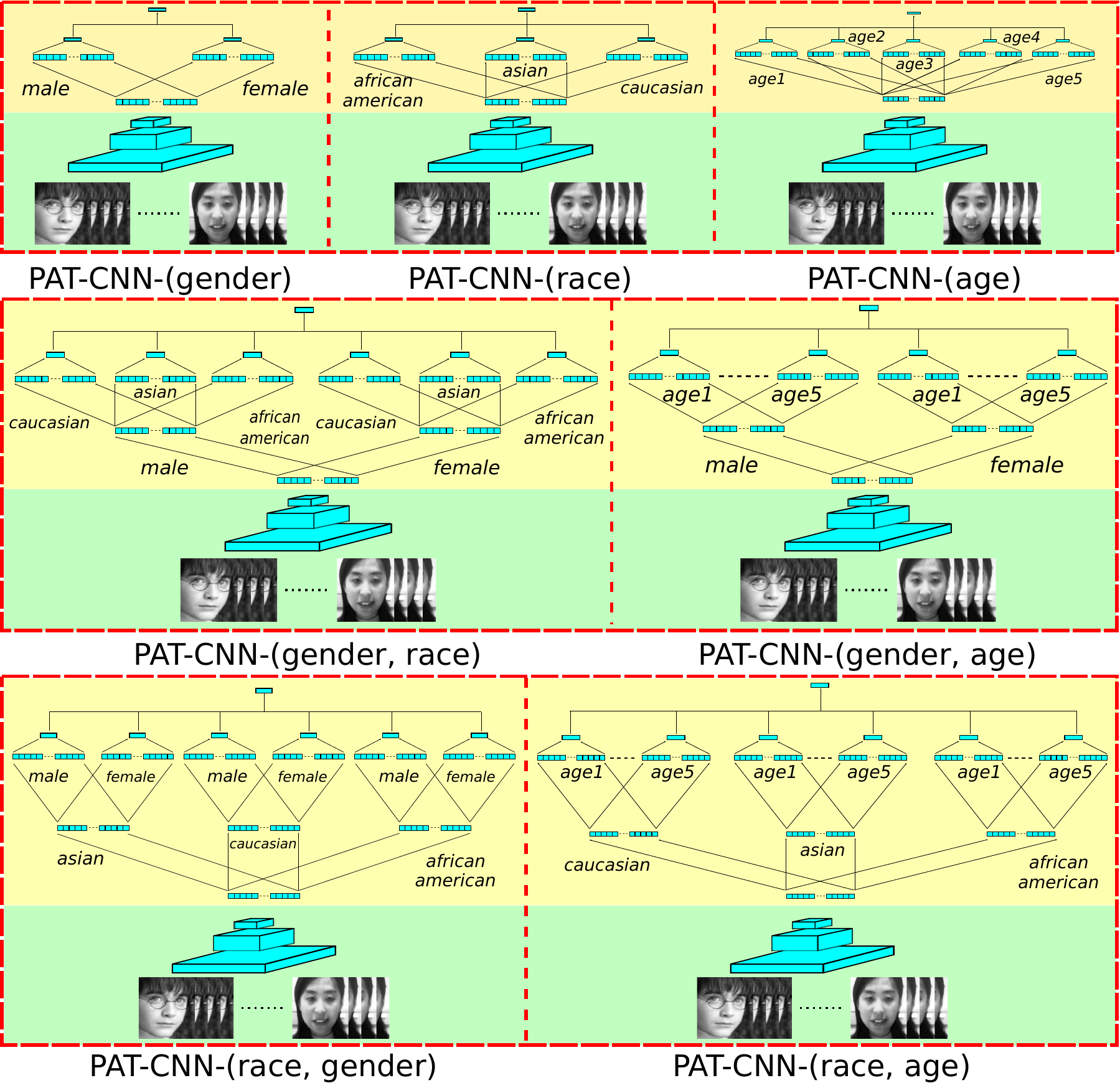}
   \caption{\small{Variants of PAT-CNNs evaluated in the experiments. Best viewed in color.}}
   \label{fig:clusteringCategory}
\end{figure}

In this work, a pre-trained VGG-Face CNN model~\cite{parkhi2015deep} and a ResNet-34 pre-trained on the ImageNet dataset~\cite{russakovsky2015imagenet} were employed as our backbone CNN networks.

\noindent  \emph{\textbf{Training baseline models:}} For experiments on the BU-3DFE, CK+, MMI, and SFEW datasets, two baseline CNNs, i.e., VGG-Face model and ResNet-34, were fine-tuned by using its own training set and the RAF-DB training set. We employed $50\%$ images from the RAF-DB in every mini-batch. For experiments on the RAF-DB dataset, they were fine-tuned by only using the RAF-DB training set.

\noindent  \emph{\textbf{Training the PAT-CNN models:}} The proposed PAT-CNN models used the same training strategy as the two baseline models, i.e., always including RAF-DB in training, for evaluation on each dataset. Note, only images from the RAF-DB dataset were used to calculate the PAT loss and all training samples were employed to calculate the expression loss as in Eq.~\ref{eq:joint_hierarchical_loss}. In this work, age, race, and gender were chosen to construct the PAT module. As shown in Fig.~\ref{fig:clusteringCategory}, we constructed several variants of PAT modules by varying the number of levels and/or the orders of attributes using the VGG-Face backbone. Then, an $l$-Level PAT module is denoted by a tuple \textit{(attribute\_1, attribute\_2, ...)}. Thereafter, variants of the proposed PAT-CNNs are denoted as \textit{PAT-VGG-(attribute\_1, attribute\_2, ...)} or \textit{PAT-ResNet-(attribute\_1, attribute\_2, ...)}.

\subsection{Experimental Results}

\subsubsection{Result analysis for the RAF-DB}
\begin{table}[th]	
\setlength{\tabcolsep}{0.1in}
  \begin{center}
    \caption{\small{Performance comparison on the RAF-DB. Some papers report performance as an average of diagonal values of confusion matrix. We convert them to regular accuracy for fair comparison.} }
    \label{tab:results_raf1}
 \scalebox{0.85}{
    \begin{tabular}{c|c}
    \hline
    Method                             &  Accuracy \\\hline \hline
    Kuo et al.~\cite{kuo2018compact}   &  72.21  \\
    FSN~\cite{zhaofeature}             &  81.10  \\
    MRE-CNN~\cite{fan2018multi}        &  82.63  \\
    baseDCNN~\cite{li2017reliable}     &  82.66 \\
    DLP-CNN~\cite{li2017reliable}      &  82.84 \\
    Center Loss~\cite{li2017reliable}  &  82.86 \\
    PG-CNN~\cite{lipatch}              &  \textbf{83.27} \\
    \hline
    VGG (baseline)                     &  81.29 \\
    \hline
    VGG(age)						       &  80.31 \\
    VGG(race)                           &  80.55 \\
    VGG(gender)                         &  80.38 \\
    \hline
    AT-VGG-(gender, race)               &  82.53 \\
    \hline
    PAT-VGG-(age)                       &  81.91 \\
    PAT-VGG-(race)                      &  82.40 \\
    PAT-VGG-(gender)                    &  82.63 \\
    PAT-VGG-(race, age)                 &  82.99 \\
    PAT-VGG-(race, gender)              &  83.44 \\
    PAT-VGG-(gender, age)               &  83.02 \\
    \textbf{PAT-VGG-(gender, race)}     &  \textbf{83.83} \\
    \hline
    ResNet (baseline)                   &  81.81 \\
    \textbf{PAT-ResNet-(gender, race)}  &  \textbf{84.19} \\
    \hline
    \end{tabular}
    \label{tab:results_raf}
    }
  \end{center}
\end{table}

The experimental results on the RAF-DB are summarized in Table~\ref{tab:results_raf}. All the variants of PAT-CNNs outperform the two baseline CNNs (VGG and ResNet), which do not consider attribute information. Furthermore, the improvement is more obvious as more attribute information is exploited, i.e., 3-level PAT-VGGs vs 2-level PAT-VGGs. Among the 2-level PAT-VGGs, PAT-VGG-(gender) achieves the best performance because the samples are more balanced in gender and the clustering performs better with gender. In contrast, the age samples are highly imbalanced, e.g., age3 (20-39) contains more than 50\% of the images. PAT-VGG-(gender, race) achieves the best results among all VGG-based methods in comparison. Both PAT-VGG-(gender, race) and PAT-ResNet-(gender, race) outperform the state-of-the-art methods evaluated on the RAF-DB. Therefore, we only reported PAT-VGG-(gender, race) and PAT-ResNet-(gender, race) in the following experiments on other four datasets.

In addition, we reported the results of three attribute-specific CNNs, i.e., VGG(age), VGG(race), and VGG(gender), which were trained from subsets of dataset as illustrated in Fig.~\ref{fig:introductionStructure} (b). As expected, the performances of the attribute-specific VGGs are consistently worse than the baseline VGG due to insufficient training data in subsets. Furthermore, we have demonstrated that the soft-clustering with probability outperforms hard-clustering, i.e., AT-VGG-(gender, race), using the same network structure.

\begin{figure}[th]
   \centering
   \includegraphics[width=0.4\textwidth]{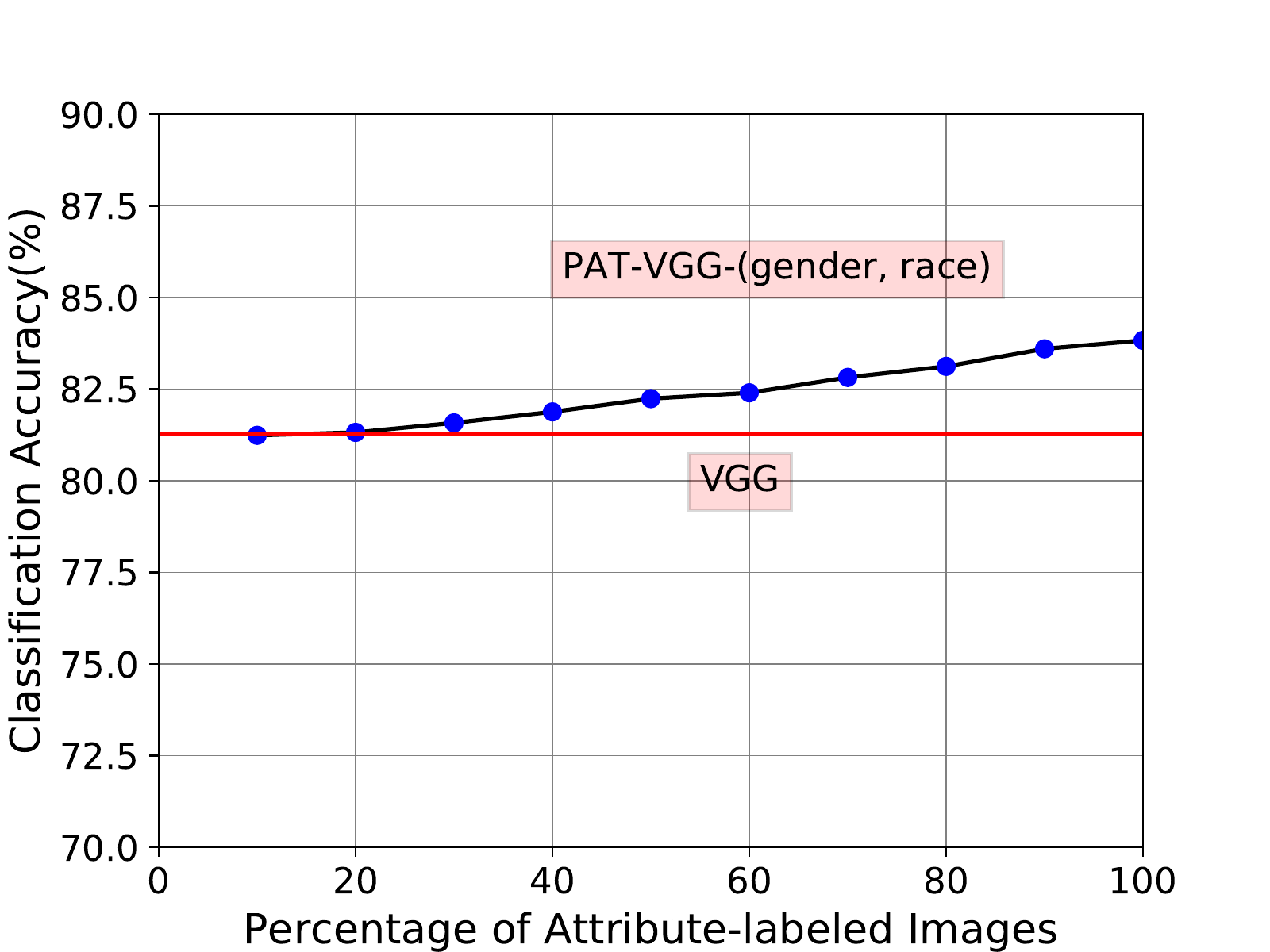}
   \caption{\small{Performance analysis by varying the percentage of attribute-labeled images in the RAF-DB dataset.}}
   \label{fig:RAF_percentage}
\end{figure}

Moreover, we evaluated the semi-supervised learning strategy for PAT-CNN training, which is critical for real-world applications, where attribute information may be missing or incomplete. Specifically, we varied the percentage of attribute-labeled images from $100\%$ to $10\%$ by randomly removing attribute labels for RAF-DB images. As shown in Fig.~\ref{fig:RAF_percentage}, the performance of PAT-VGG-(gender, race) beats that of the VGG baseline when more than $20\%$ of RAF-DB images have attribute labels.

\begin{table}[th!]
  \begin{center}
  	\caption{\small{Performance comparison on the BU-3DFE.}}
	\scalebox{0.75}{
    \begin{tabular}{c|c|c|c}
    \hline
    Method                                 &  Accuracy              &  Classes       &    Feature     \\\hline \hline
    Wang et al.~\cite{wang20063d}          &    61.79               &  6             &    3D-Static        \\
    Berretti et al.~\cite{berretti2010set} &    77.54               &  6             &    3D-Static        \\
    Yang et al.~\cite{yang2015automatic}   &    \textbf{84.80}      &  6             &    3D-Static        \\
    Lopes et al.~\cite{lopes2017facial}    &    72.89               &  6             &    2D-Static        \\
    Lai et al.~\cite{lai2018emotion}       &    74.25               &  6             &    2D-Static        \\
    IA-gen~\cite{yang2018identity}         &    76.83               &  6             &    2D-Static        \\
    DPND~\cite{chen2018deep}               &    78.4                &  6             &    2D-Static        \\
    Zhang et al.~\cite{zhang2018joint}     &    80.95               &  6             &    2D-Static        \\
    DeRL~\cite{yang2018facial}             &    84.17               &  6             &    2D-Static        \\
    \hline
    VGG (baseline)                         &    81.08               &     6          &  2D-Static       \\
    \textbf{PAT-VGG-(gender, race)}        &    \textbf{83.58}      &     6          &  2D-Static       \\
    \hline
    ResNet (baseline)                      &    81.31               &     6          &  2D-Static       \\
    \textbf{PAT-ResNet-(gender, race)}     &    \textbf{83.17}      &     6          &  2D-Static       \\
    \hline
    \end{tabular}}
    \label{tab:results_bu3dfe}
  \end{center}
\end{table}

\begin{table}[th!]
  \begin{center}
  	\caption{\small{Performance comparison on the CK+.}}
	\scalebox{0.75}{
    \begin{tabular}{c|c|c|c}
    \hline
    Method                                     &  Accuracy          &  Classes  &  Feature     \\\hline \hline
    TMS~\cite{jain2011facial}                  &   91.89            &    6      & Dynamic      \\
    Inception~\cite{mollahosseini2016going}    &   93.2             &    6      & Static      \\
    DLP-CNN~\cite{li2017reliable}              &   95.78            &    6      & Static      \\
    Lopes et al.~\cite{lopes2017facial}        &   96.76            &    6      & Static      \\
    PPDN~\cite{zhao2016peak}     	       &   \textbf{97.3}    &    6      & Static      \\
    \hline
    VGG (baseline)                             &    93.42           &     6          &  Static       \\
    \textbf{PAT-VGG-(gender, race)}            &    \textbf{95.58}  &     6          &  Static       \\
    \hline
    ResNet (baseline)                          &    93.31           &     6          &  Static       \\
    \textbf{PAT-ResNet-(gender, race)}         &    \textbf{96.01}  &     6          &  Static       \\
    \hline
    \hline
    3DCNN~\cite{liu2014deeplyaccv}             &   85.9             &    7      & Dynamic      \\
    ITBN~\cite{WangZH2013}                     &   86.3             &    7      & Dynamic      \\
    F-Bases~\cite{sariyanidi2017learning}      &   89.01            &    7      & Dynamic      \\
    Cov3D~\cite{sanin2013spatio}               &   92.3             &    7      & Dynamic      \\
    3DCNN-DAP~\cite{liu2014deeplyaccv}         &   92.4             &    7      & Dynamic      \\
    STM-ExpLet~\cite{liu2014learning}          &   94.19            &    7      & Dynamic      \\
    LOMo~\cite{sikka2016lomo}                  &   95.1             &    7      & Dynamic      \\
    STM~\cite{chu2016selective}                &   96.40            &    7      & Dynamic      \\
    DTAGN~\cite{jung2015joint}                 &   97.25            &    7      & Dynamic     \\
    Center Loss~\cite{wen2016discriminative}   &   92.25            &    7      & Static      \\
    Island Loss~\cite{cai2018island}           &   94.39            &    7      & Static      \\
    IA-gen~\cite{yang2018identity}             &   96.57            &    7      & Static      \\
    IACNN~\cite{meng2017identity}              &   95.37            &    7      & Static      \\
    DeRL~\cite{yang2018facial}                 &   \textbf{97.3}    &    7      & Static      \\
    \hline
    VGG (baseline)                             &    92.97           &     7          &  Static       \\
    \textbf{PAT-VGG-(gender, race)}            &    \textbf{95.31}  &     7          &  Static       \\
    \hline
    ResNet (baseline)                          &    92.56           &     7          &  Static       \\
    \textbf{PAT-ResNet-(gender, race)}         &    \textbf{95.82}  &     7          &  Static       \\
    \hline
    \end{tabular}}
    \label{tab:results_ckplus}
  \end{center}
\end{table}

\begin{table}[th!]
  \begin{center}
  	\caption{\small{Performance comparison on the MMI.}}
	\scalebox{0.75}{
    \begin{tabular}{c|c|c|c}
    \hline
    Method                                   &  Accuracy           &  Classes    &  Feature   \\\hline \hline
    3DCNN~\cite{liu2014deeplyaccv}           &   53.2              &    6       &  Dynamic    \\
    ITBN~\cite{WangZH2013}                   &   59.7              &    6       &  Dynamic    \\
    3DCNN-DAP~\cite{liu2014deeplyaccv}       &   63.4              &    6       &  Dynamic    \\
    DTAGN~\cite{jung2015joint}               &   70.24             &    6       &  Dynamic    \\
    F-Bases~\cite{sariyanidi2017learning}    &   73.66             &    6       &  Dynamic    \\
    STM-ExpLet~\cite{liu2014learning}        &   \textbf{75.12}    &    6       &  Dynamic    \\
    IACNN~\cite{meng2017identity}            &   71.55             &    6       &  Static     \\
    DeRL~\cite{yang2018facial}               &   73.23             &    6       &  Static     \\
    Center Loss~\cite{wen2016discriminative} &   73.40             &    6       &  Static     \\
    Island Loss~\cite{cai2018island}         &   74.68             &    6       &  Static     \\
    \hline
    VGG (baseline)                           &    69.39            &     6          &  Static       \\
    \textbf{PAT-VGG-(gender, race)}          &    \textbf{73.56}   &     6          &  Static       \\
    \hline
    ResNet (baseline)                                   &    70.99            &     6          &  Static       \\
    \textbf{PAT-ResNet-(gender, race)}       &    \textbf{74.04}   &     6          &  Static       \\
    \hline
    \end{tabular}}
    \label{tab:results_mmi}
  \end{center}
\end{table}

\subsection{Result analysis for the three posed datasets}

As shown in Table~\ref{tab:results_bu3dfe}, ~\ref{tab:results_ckplus}, and ~\ref{tab:results_mmi}, the proposed PAT-CNNs outperforms the baseline CNNs for both backbone structures and also achieve comparable results as the state-of-the-art methods evaluated on the three posed datasets. Note that most of the state-of-the-art methods utilized dynamic features extracted from image sequences, while the proposed PAT-CNN is trained on static images, which is more favorable for online applications or snapshots. Yang et al.~\cite{yang2015automatic} achieved the highest performance on the BU-3DFE dataset by employing geometric features of the 3D shape model. Although the cGAN-based methods (DeRL~\cite{yang2018facial} and IA-gen~\cite{yang2018identity}) achieved high performance on the BU-3DFE and CK+ datasets, they are not end-to-end systems and also require higher computational cost. We are aware that PPDN~\cite{zhao2016peak} also has the best performance on the CK+ dataset owing to utilizing neutral images as reference. Island Loss~\cite{cai2018island} achieved the best performance on the MMI dataset by utilizing an average fusion of the three images from the same sequence.

\subsection{Result analysis for the SFEW dataset}

More importantly, the proposed PAT-CNN was also evaluated on the SFEW dataset, which contains unconstrained and thus, more natural facial expressions and has been used as a benchmark to evaluate facial expression recognition systems in the wild. Note that the top three methods reported on the SFEW testing set~\cite{Kim2015hierarchical,yu2015image,cai2018island} utilized an ensemble of CNNs. As shown in Table~\ref{tab:results_sfew}, the proposed PAT-CNNs beat the baseline CNNs for both validation set and testing set by a large margin. More impressively, the proposed PAT-CNNs with both backbone structures using a single model achieve the best performance on the testing set among all the methods compared with.

\begin{table}[th]	
  \begin{center}
    \caption{\small{Performance comparison on the SFEW.} }
    	\scalebox{0.85}{
    \begin{tabular}{c|c|c|c}
    \hline
    Method                                  &  Model        & Val             & Test\\    \hline \hline
    Kim et al.~\cite{Kim2015hierarchical}   &   Ensemble    & 53.9            & \textbf{61.6}\\
    Yu et al.~\cite{yu2015image}            &   Ensemble    & \textbf{55.96}  & 61.29 \\
    Island Loss~\cite{cai2018island}        &   Ensemble    & 52.52           & 59.41 \\
    \hline
    Island Loss~\cite{cai2018island}        &   Single      & 51.83           & 56.99 \\
    Ng et al.~\cite{Ng2015deep}             &   Single      & 48.5            & 55.6  \\
    Yao et al.~\cite{yao2015capturing}      &   Single      & 43.58           & 55.38 \\
    IACNN~\cite{meng2017identity}           &   Single      & 50.98           & 54.30 \\
    Center Loss~\cite{wen2016discriminative}&   Single      & 48.85           & 53.76 \\
    Sun et al.~\cite{sun2015combining}      &   Single      & 51.02           & 51.08 \\
    STTLDA~\cite{zong2016emotion}           &   Single      & --              & 50    \\
    Kaya et al.~\cite{kaya2015contrasting}  &   Single      & 53.06           & 49.46 \\
    baseline of SFEW~\cite{dhall2015video}  &   Single      & 35.93           & 39.13 \\
    FN2EN~\cite{ding2017facenet2expnet}     &   Single      & 55.15           & --   \\
    DLP-CNN~\cite{li2017reliable}           &   Single      & 51.05           & --   \\
    Inception~\cite{mollahosseini2016going} &   Single      & 47.7            & --   \\
    \hline
    VGG (baseline)                          &   Single     &  49.77          & 57.26           \\
    \textbf{PAT-VGG-(gender, race)}         &   Single     &  \textbf{53.21} & \textbf{62.90}  \\
    \hline
    ResNet (baseline)                       &   Single     &  50.00          & 56.72           \\
    \textbf{PAT-ResNet-(gender, race)}      &   Single     &  \textbf{52.75} & \textbf{62.10}  \\
    \hline
    \end{tabular}}
    \label{tab:results_sfew}
  \end{center}
\end{table}

\section{CONCLUSION}
In this work, we proposed a novel PAT-CNN along with a forward-backward propagation algorithm to learn expression-related features in a hierarchical structure by explicitly modeling identity-related attributes in the CNN. Our work differs from the other unsupervised clustering methods in that the proposed PAT-CNN is capable of building semantically-meaningful clusters from which expression-related features are learned to alleviate the inter-subject variations. Furthermore, data samples have different weights, i.e., the probabilities of being assigned to the nodes, in training the PAT nodes; and the probabilities are also used as the weights to build a strong classifier for expression recognition. More importantly, most of the existing facial expression datasets do not have attribute labels. The proposed loss function enables a semi-supervised learning strategy to improve expression recognition for these databases with the help of additional attribute-labeled data, such as the RAF-DB.

Facial expression recognition has achieved great progress on posed facial displays in lab settings (Table~\ref{tab:results_bu3dfe}, ~\ref{tab:results_ckplus}, and ~\ref{tab:results_mmi}), but suffers in real world conditions (Table~\ref{tab:results_raf} and~\ref{tab:results_sfew}), which significantly impedes its applications. The PAT-CNNs achieved the best performance on both spontaneous databases (RAF-DB and SFEW datasets) in the real-world setting by explicitly handling large attribute variations. In the future, we plan to apply the PAT-CNN to other classification problems that suffer from large intra-class variations.


\section{Acknowledgement}
This work is supported by National Science Foundation under CAREER Award IIS-1149787.

{
\small
\bibliographystyle{ieee.bst}
\bibliography{../../../bibliography/abbrev_short,../../../bibliography/machine_learning/ty-literature_graphical_model,../../../bibliography/ty-literature_misc,../../../bibliography/ty-literature_self,../../../bibliography/emotion/ty-literature_AU_Exp_Emotion_rec,../../../bibliography/ty-literature_audiovisual_ASR,../../../bibliography/ty-literature_facial_feature_detect_track,../../../bibliography/machine_learning/ty-literature_unsupervised_feature_learning,../../../bibliography/machine_learning/ty-literature_gan,../../../bibliography/ty-literature_database,../../../bibliography/machine_learning/ty-literature_machine_learning,../../../bibliography/machine_learning/ty-literature_deep_learning,../../../bibliography/machine_learning/ty-literature_metric_learning,../../../bibliography/ty-literature_statstical_models_alignment,../../../bibliography/object_classification/ty-literature_object_detection,../../../bibliography/ty-literature_psychology,../../../bibliography/machine_learning/ty-literature_loss,../../../bibliography/machine_learning/ty-literature_multi_task_learning,../../../bibliography/emotion/ty-literature_gan_exp.bib,../../../bibliography/ty-literature_EmotiW.bib,../../../bibliography/machine_learning/ty-literature_cnn_clustering}
}

\end{document}